\documentclass[letterpaper, 10 pt, conference]{ieeeconf}

\IEEEoverridecommandlockouts       

\makeatletter
\let\NAT@parse\undefined
\makeatother
\usepackage[numbers]{natbib}

\usepackage{multicol}
\usepackage[bookmarks=true]{hyperref}

\usepackage[usenames,dvipsnames,svgnames]{xcolor}
\usepackage{times} 
\usepackage{amsmath} 
\usepackage{amssymb}  
\usepackage{algorithm}
\usepackage[noend]{algpseudocode}
\usepackage{comment}
\usepackage{graphicx}
\usepackage{subfigure}
\usepackage{tikz}
\usepackage{etoolbox}
\usepackage{xspace}
\usepackage[letterspace=0,tracking=true]{microtype}
\usepackage{comment}
\usepackage[utf8]{inputenc}
\usepackage[T2A,T1]{fontenc}
\usepackage{booktabs}
\usepackage{paralist}
\usepackage{cleveref}
\usepackage[export]{adjustbox}
\usepackage{scalerel}

\graphicspath{{images/}}

\definecolor{Red}{rgb}{1,0,0}
\definecolor{Green}{rgb}{0,0.69,0}
\definecolor{Blue}{rgb}{0,0,1}
\definecolor{LightBlue}{rgb}{0,0.5,1}
\definecolor{veryLightBlue}{rgb}{0.85,0.98,1}
\definecolor{veryLightGreen}{rgb}{0.6,1,0.6}
\definecolor{Skin}{rgb}{1,0.71,0.69}
\definecolor{Grey}{rgb}{0.5,0.5,0.5}
\definecolor{LightGrey}{rgb}{0.6,0.6,0.6}
\definecolor{Black}{rgb}{0,0,0}
\definecolor{White}{rgb}{1,1,1}
\newcommand{\red}{\color{Red}}
\newcommand{\green}{\color{Green}}
\newcommand{\blue}{\color{Blue}}

\newcommand{\orange}{\color{Bittersweet}}

\newcommand{\purple}{\color{purple}}

\newbool{blind}
\setbool{blind}{false}

\ifbool{blind}
{\AtBeginEnvironment{blind}{\comment}%
  \AtEndEnvironment{blind}{\endcomment}}{}

\makeatletter
\DeclareRobustCommand\onedot{\futurelet\@let@token\@onedot}
\def\@onedot{\ifx\@let@token.\else.\null\fi\xspace}

\def\etal{\emph{et al}\onedot}
\newcommand{\eg}{e.g.,\xspace}

\newcommand{\ie}{i.e.,\xspace}

\newcommand{\vs}{vs\onedot\xspace}
\makeatother

\usetikzlibrary{calc}
\usetikzlibrary{fit}
\usetikzlibrary{positioning}
\usetikzlibrary{decorations.markings}
\usetikzlibrary{shapes.geometric}
\usetikzlibrary{shapes.multipart}
\usetikzlibrary{bayesnet}
\usetikzlibrary{tikzmark}
\usetikzlibrary{backgrounds} 
\usetikzlibrary{shapes}
\usetikzlibrary{3d}
\usetikzlibrary{arrows.meta}
\usetikzlibrary{shapes.geometric}

\definecolor{pinegreen}{cmyk}{0.92,0,0.59,0.25}
\definecolor{royalblue}{cmyk}{1,0.50,0,0}
\tikzstyle{cblue}=[circle, draw, thin,fill=cyan!20, scale=0.8]
\tikzstyle{obs}=[circle, draw, thin,fill=gray!20, scale=0.8]
\tikzstyle{qgre}=[circle, draw, scale=0.8]
\tikzstyle{rpath}=[thick, black, opacity=0.4]

\tikzstyle{pathnode}=[circle, draw, scale=0.04]
\tikzstyle{graphnode}=[circle, draw, scale=0.15,ultra thin]
\tikzstyle{graphedge}=[-{Latex[black,length=0.15ex,width=0.15ex]}, shorten >= 0.04ex, ultra thin]
\tikzstyle{graphdashed}=[dash pattern=on 0.2pt off 0.1pt]

\newcommand{\bugtrap}%
{
  \draw[ultra thin, purple] (0,0,0)--(0,1,0) (0,1,0)--(1,1,0) (1,1,0)--(1,0,0) (1,0,0)--(0,0,0);
  \coordinate (a) at (0.25,0.25,0);
  \coordinate (b) at ($(a)+(0.21,0,0)$);
  \coordinate (c) at ($(b)+(0,0.24,0)$); 
  \coordinate (d) at ($(a)+(0,0.5,0)$); 
  \coordinate (e) at ($(d)+(0.5,0,0)$); 
  \coordinate (f) at ($(e)+(0,-0.5,0)$); 
  \coordinate (g) at ($(f)+(-0.21,0,0)$); 
  \coordinate (h) at ($(g)+(0,0.24,0)$);
  \draw[ultra thin, purple] (a)--(b) (b)--(c) (d)--(a) (d)--(e) (e)--(f) (f)--(g) (g)--(h);
}

\newcommand{\plannerrnn}[4]%
{
  \node[graphnode, above=1 of start, xshift=-10ex, xshift=#2, yshift=0ex, yshift=#3, #4] (gstartX#1) {};
  \node[graphnode, above=1 of n1, xshift=-1ex, xshift=#2, yshift=0ex, yshift=#3, #4] (gn1X#1) {};
  \node[graphnode, above=1 of n2, xshift= 1ex, xshift=#2, yshift=5ex, yshift=#3, #4] (gn2X#1) {};
  \node[graphnode, above=1 of n3, xshift=-1ex, xshift=#2, yshift=-5ex, yshift=#3, #4] (gn3X#1) {};
  \node[graphnode, above=1 of n4, xshift=2ex, xshift=#2, yshift=0ex, yshift=#3, #4] (gn4X#1) {};
  \node[graphnode, above=1 of n5, xshift=8ex, xshift=#2, yshift=5ex, yshift=#3, #4] (gn5X#1) {};
  \node[graphnode, above=1 of n6, xshift=10ex, xshift=#2, yshift=-5ex, yshift=#3, #4] (gn6X#1) {};

  \draw[graphedge, graphdashed] (gstartX#1) -- (start);
  \draw[graphedge, graphdashed] (gn1X#1) -- (n1);
  \draw[graphedge, graphdashed] (gn2X#1) -- (n2);
  \draw[graphedge, graphdashed] (gn3X#1) -- (n3);
  \draw[graphedge, graphdashed] (gn4X#1) -- (n4);
  \draw[graphedge, graphdashed] (gn5X#1) -- (n5);
  \draw[graphedge, graphdashed] (gn6X#1) -- (n6);
  \draw[graphedge, graphdashed] (gn6X#1) -- (n7);
  \draw[graphedge] (gstartX#1) -- (gn1X#1);
  \draw[graphedge] (gstartX#1) -- (gn2X#1);
  \draw[graphedge] (gn1X#1) -- (gn3X#1);
  \draw[graphedge] (gn1X#1) -- (gn4X#1);
  \draw[graphedge] (gn4X#1) -- (gn5X#1);
  \draw[graphedge] (gn4X#1) -- (gn6X#1);
}

\def\ltlX{\mathbin{\scalerel*{\bigcirc}{\forall}\hspace{0.1ex}}}
\def\ltlnext{\ltlX}
\def\ltlF{\mathbin{\scalerel*{\square}{\forall}\hspace{0.1ex}}}
\def\ltlfinally{\ltlF}
\def\ltlG{\mathbin{\scalerel*{\lozenge}{\forall}\hspace{0.1ex}}}
\def\ltlglobally{\ltlG}

\begin{document}


\title{\LARGE \bf
Encoding formulas as deep networks:\\Reinforcement learning for zero-shot execution of LTL formulas}


\ifbool{blind}{
  \author{Author Names Omitted for Anonymous Review.}
  }{%
  \author{Yen-Ling Kuo, Boris Katz, and Andrei Barbu%
    \thanks{This work was supported by the Center for Brains, Minds and Machines, NSF STC award 1231216, the Toyota Research Institute, the DARPA GAILA program, the ONR Award No.N00014-20-1-2589, and the CBMM-Siemens Graduate Fellowship.}%
    \thanks{\protect\raggedright CSAIL and CBMM, MIT
      {\tt\small \{ylkuo,boris,abarbu\}@mit.edu}}%
  }}

\maketitle
\thispagestyle{empty}
\pagestyle{empty}

\begin{abstract}
  We demonstrate a reinforcement learning agent which uses a compositional
  recurrent neural network that takes as input an LTL formula and determines
  satisfying actions. The input LTL formulas have never been seen before, yet
  the network performs zero-shot generalization to satisfy them. This is a novel
  form of multi-task learning for RL agents where agents learn from one diverse
  set of tasks and generalize to a new set of diverse tasks. The formulation of
  the network enables this capacity to generalize. We demonstrate this ability
  in two domains. In a symbolic domain, the agent finds a sequence of letters
  that is accepted. In a Minecraft-like environment, the agent finds a sequence
  of actions that conform to the formula. While prior work could learn to
  execute one formula reliably given examples of that formula, we demonstrate
  how to encode all formulas reliably. This could form the basis of new
  multi-task agents that discover sub-tasks and execute them without any
  additional training, as well as the agents which follow more complex
  linguistic commands. The structures required for this generalization are
  specific to LTL formulas, which opens up an interesting theoretical question:
  what structures are required in neural networks for zero-shot generalization
  to different logics?
\end{abstract}

\IEEEpeerreviewmaketitle

\section{Introduction}

\begin{figure}
  \centering
  \vspace{0.5ex}
  \begin{normalsize}
    \scalebox{1.1}{\begin{tikzpicture}
      \node[draw,solid,black] (frame) {\includegraphics[width=0.1\textwidth]{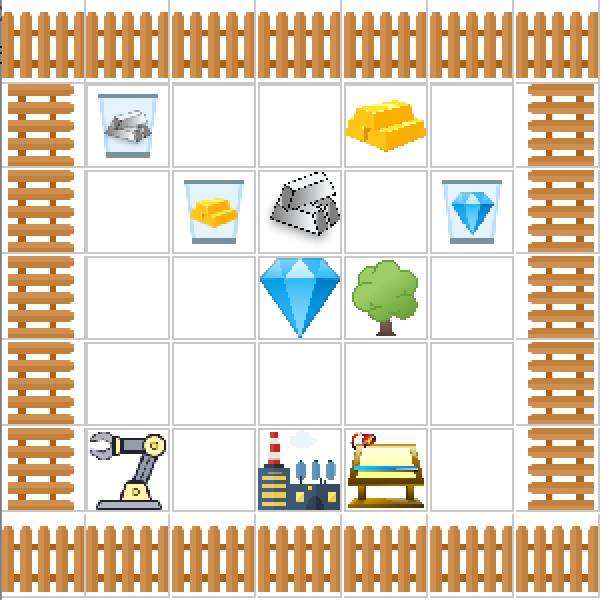}};
      \node[above=0.5 of frame] (encoder) {Feature extraction network};
      \node[Black,above=0.3 of encoder,xshift=-12.5ex] (factory) {gem};
      \node[Black, right=3 of factory] (gem) {factory};
      
      \node[Black, above=1 of factory] (next) {$\ltlglobally$};
      \node[Black, above=1 of gem] (globally) {$\ltlfinally$};
      \node (mid) at ($(next)!0.5!(globally)$) {};
      \node[Black, above=1 of mid] (and) {$\wedge$};
      \node[Black, above=1 of and] (action) {\emph{Action}};

      \draw[-{Latex[length=1.2ex,width=1.2ex]},thin,black] (frame) -- (encoder);
      \draw[-{Latex[length=1.2ex,width=1.2ex]},thin,purple,dotted] (encoder) -- (gem);
      \draw[-{Latex[length=1.2ex,width=1.2ex]},thin,purple,dotted] (encoder) -- (factory);
      \draw[-{Latex[length=1.2ex,width=1.2ex]},thin,purple,dotted] (encoder) -- (globally);
      \draw[-{Latex[length=1.2ex,width=1.2ex]},thin,purple,dotted] (encoder) -- (next);
      \draw[-{Latex[length=1.2ex,width=1.2ex]},thin,purple,dotted] (encoder) -- (and);

      \draw[thin,orange,densely dashed,-{Latex[length=1.2ex,width=1.2ex]}] ($(factory.west)+(-1,0.05)$) -- ($(factory.west)+(0,0.05)$);
      \draw[thin,orange,densely dashed,-{Latex[length=1.2ex,width=1.2ex]}] ($(factory.east)+(0,0.05)$) -- ($(factory.east)+(1,0.05)$);
      \draw[{Bar[width=3ex].Butt Cap[sep=0.5ex]}-{Latex[length=1.2ex,width=1.2ex]},thin,Green]
      ($(factory.north east)+(0.1,-0.1)$) to[out=90,in=-30] (next);
      \draw[{Bar[width=3ex].Butt Cap[sep=0.5ex]}-{Latex[length=1.2ex,width=1.2ex]},thin,Blue]
      ($(next.south west)+(-0.43,0.15)$) to[out=-90,in=140] (factory);
      
      \draw[thin,orange,densely dashed,-{Latex[length=1.2ex,width=1.2ex]}] ($(gem.west)+(-1,0.05)$) -- ($(gem.west)+(0,0.05)$);
      \draw[thin,orange,densely dashed,-{Latex[length=1.2ex,width=1.2ex]}] ($(gem.east)+(0,0.05)$) -- ($(gem.east)+(1,0.05)$);
      \draw[{Bar[width=3ex].Butt Cap[sep=0.5ex]}-{Latex[length=1.2ex,width=1.2ex]},thin,Green]
      ($(gem.north east)+(0.1,-0.1)$) to[out=90,in=-30] (globally);
      \draw[{Bar[width=3ex].Butt Cap[sep=0.5ex]}-{Latex[length=1.2ex,width=1.2ex]},thin,Blue]
      ($(globally.south west)+(-0.43,0.15)$) to[out=-90,in=140] (gem);

      \draw[thin,orange,densely dashed,-{Latex[length=1.2ex,width=1.2ex]}] ($(next.west)+(-1,0.05)$) -- ($(next.west)+(0,0.05)$);
      \draw[thin,orange,densely dashed,-{Latex[length=1.2ex,width=1.2ex]}] ($(next.east)+(0,0.05)$) -- ($(next.east)+(1,0.05)$);
      \draw[{Bar[width=3ex].Butt Cap[sep=0.5ex]}-{Latex[length=1.2ex,width=1.2ex]},thin,Green]
      ($(next.north east)+(0.1,-0.1)$) to[out=90,in=-140] (and);
      \draw[{Bar[width=3ex].Butt Cap[sep=0.5ex]}-{Latex[length=1.2ex,width=1.2ex]},thin,Blue]
      ($(and.south west)+(-0.43,0.15)$) to[out=-90,in=110] (next);

      \draw[thin,orange,densely dashed,-{Latex[length=1.2ex,width=1.2ex]}] ($(globally.west)+(-1,0.05)$) -- ($(globally.west)+(0,0.05)$);
      \draw[thin,orange,densely dashed,-{Latex[length=1.2ex,width=1.2ex]}] ($(globally.east)+(0,0.05)$) -- ($(globally.east)+(1,0.05)$);
      \draw[{Bar[width=3ex].Butt Cap[sep=0.5ex]}-{Latex[length=1.2ex,width=1.2ex]},thin,Green]
      ($(globally.north east)+(0.1,-0.1)$) to[out=90,in=-120] (and);
      \draw[{Bar[width=3ex].Butt Cap[sep=0.5ex]}-{Latex[length=1.2ex,width=1.2ex]},thin,Blue]
      ($(and.south west)+(-0.43,0.15)$) to[out=-90,in=145] (globally);

      \draw[thin,orange,densely dashed,-{Latex[length=1.2ex,width=1.2ex]}] ($(and.west)+(-1,0.05)$) -- ($(and.west)+(0,0.05)$);
      \draw[thin,orange,densely dashed,-{Latex[length=1.2ex,width=1.2ex]}] ($(and.east)+(0,0.05)$) -- ($(and.east)+(1,0.05)$);

      \draw[-{Latex[length=1.2ex,width=1.2ex]},thin,Red] (and) to (action);
    \end{tikzpicture}}\\[2ex]
  \end{normalsize}
  \caption{A structured compositional deep network encoding one formula,
    $\protect\ltlglobally\textsc{gem}\wedge\protect\ltlfinally\textsc{factory}$. This
    formula corresponds to a command such as ``Hold the gem and always get to the
    factory''. Each operator and predicate in the formula is represented by a
    network, shown in black, selected from a trained collection of
    sub-networks. Sub-networks are {\orange RNNs} which maintain state over
    time, shown in {\orange orange}. Each sub-network takes as input {\purple
      features} extracted from the surroundings of the robot using a co-trained
    network, shown in {\purple dotted purple}. The {\green next state} of each
    sub-network is decoded by a {\green linear layer} and passed to its parents,
    shown in {\green green}. The {\blue previous state} of each sub-network is
    decoded using a {\blue linear layer} and passed to its children, shown in
    {\blue blue}. Finally, the state of the root node is decoded into a
    distribution over the value of the actions the robot can take at the current
    time step. Crucially, due to the compositional representation employed,
    novel formulas that were never seen at training time can be encoded and
    followed on novel never-before-seen maps. This couples the power of deep
    networks to learn to extract features and engage in complex behaviors with
    knowledge about the structure of formulas, to perform zero-shot
    execution. See \protect\cref{fig:model-execution} for an example of the
    model in action.}
  \label{fig:model}
\end{figure}

Robots must execute commands that are extended in time while being responsive to
changes in their environments.
A popular representation for such commands is linear temporal logic, LTL~\cite{pnueli1977temporal}.
Commands expressed in LTL encode both spatial and temporal constraints that
should be true while executing the command.
Executing such commands is particularly difficult in robotics because
integration is required between the complex symbolic reasoning that finds
satisfying sequences of moves for an LTL command and data-driven perceptual
capabilities required to sense the environment.
While individual formulas can be learned by deep networks with extensive
experience, we demonstrate how to compose together tasks and skills to learn a
general principle of how to encode all LTL formulas and follow them without
per-formula experience.
We demonstrate how to integrate the learning abilities of neural networks with
the symbolic structure of LTL commands to achieve a new capability: learning to
perform end-to-end zero-shot execution of LTL commands.

Given a command represented as an LTL formula, our approach turns that formula
into a specific recurrent deep network which encodes the meaning of that
command; see~\cref{fig:model}.
The resulting network takes as input the current map state, extracts the
environment around the robot, processes it with a co-trained feature extraction network, and predicts
which actions will satisfy the formula.
This compositional approach ties together neural networks and symbolic reasoning
allowing any LTL formula to be encoded and followed, even if it has never been
seen before at training time.

\begin{figure}
  \centering
  \vspace{0.5ex}
  \begin{normalsize}
    $\ltlglobally\textsc{gem}\wedge\ltlfinally\textsc{factory}$\\[2ex]
  \begin{tikzpicture}
      \node[anchor=south west,inner sep=0] at (0,0) {\includegraphics[width=0.4\textwidth]{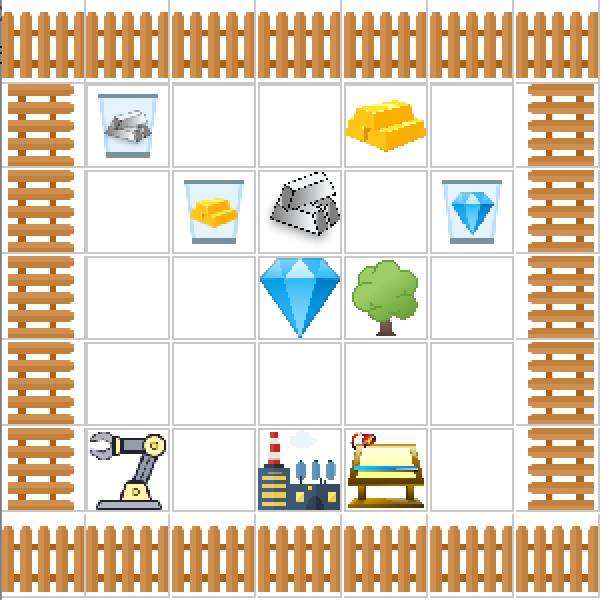}};
      \draw[dashed,red,ultra thick,rounded corners,->] (1.5,1.5) -- (1.5,3.5) -- (3.5,3.5);
      \draw[dashed,red,ultra thick,rounded corners,->] (3.5,3.5) -- (3.5,1.5);
    \end{tikzpicture}
  \end{normalsize}
  \caption{The model from \protect\cref{fig:model} executing a novel command
    that was not shown at training time on a novel map. Maps and formulas are
    randomly generated. In this case the robot executes
    $\protect\ltlglobally\textsc{gem}\wedge\protect\ltlfinally\textsc{factory}$ 
    by going to the
    gem, picking it up, and, while holding on to the gem, going to the factory.}
  \label{fig:model-execution}
\end{figure}

In our experiments, we generate random LTL formulas and train an RL agent to
follow those formulas.
We develop a mechanism for generating hard and diverse LTL formulas, as random
instances tend to be homogeneous and trivially solved.
This is generally useful for other large-scale experiments on following commands
that can be encoded as LTL formulas.
In two different domains, \texttt{Symbol} and \texttt{Craft}, we show that this approach can learn
to execute never-before-seen formulas.
The \texttt{Symbol} domain is more akin to boolean satisfiability, where an
accepting string must be generated for an LTL formula.
The \texttt{Craft} domain is a simplified Minecraft introduced by Andreas \etal 2017
\cite{andreas2017modular} to test the integration with robotics; see
\cref{fig:model-execution} for an example of the network in \cref{fig:model}
executing a command in the \texttt{Craft} domain.
In all cases, we compare against baselines to demonstrate that each part of our
model plays a key role in encoding temporal structures.
All components of our networks are learned end-to-end, in a process that
automatically isolates the meaning of each sub-network allowing us to compose
sub-networks together in novel ways.

This work makes four contributions:
\begin{compactenum}[1.]
\item a trained deep network for following LTL commands,
\item end-to-end zero-shot execution of LTL commands,
\item co-training the feature extraction with the LTL controller,
\item an investigation of the properties of random LTL formulas.
\end{compactenum}
This work can be seen as a novel approach to composing the policies of
multi-task reinforcement learning agents in a principled manner according to a
particular logic.
While we only discuss LTL here, this approach suggests how other logics might
similarly be encoded to create new powerful zero-shot deep approaches to
reinforcement learning.
The best aspects of symbolic reasoning in robotics are compatible with deep
networks when both are correctly formulated.
Perhaps in the future, such approaches could be used for model checking with LTL
formulas.

%
%




\section{Related Work}

The most related work to what we present here is in the area of multi-task
learning and task composition.
Previous work has learned finite state machines in conjunction with robot
controllers~\cite{Araki2019learning}. The focus in that prior work is on
training extensively with one LTL formula and then following that formula; here
we show how to zero-shot follow novel LTL formulas on new maps.

A related line of research shows how to accelerate learning of new LTL formulas
by shaping rewards according to the structure of the
formulas~\cite{Camacho2019ltl}. Here we provide no example of novel formulas
whereas this prior work requires tens of thousands to millions of training
steps, in the \texttt{Craft} domain, for each new formula.

Compositions of LTL sub-formulas have been investigated
before~\cite{Sahni2017learning}. Sahni \etal (2017) show how to compose together
controllers for LTL formulas. The formulas considered are small: the largest is
well below the mean size of our formulas. Only 4 formulas are tested
thoroughly. Some zero-shot generalization is achieved to 4 formulas which have
the same structure as those in the training set or which are carefully stitched
together by hand given knowledge of their semantics. No general algorithm is
given for how to encode an arbitrary LTL formula, nor is it possible to
automatically train on a set of generated formulas.

LTL has been recently used to decompose tasks into sub-tasks, learn policies for
each of the subtasks, and to improve the reliability and generalization capabilities
of robots~\cite{Toro2018teaching}. Their approach must re-learn each subtask; in
essence, the structure of the LTL formulas themselves plays no role. Here we
show a more extreme approach where tasks encoded in LTL and composed together
and directly followed without any task-specific training. Moreover, only 10
tasks which are paired with training data are considered, whereas here we
execute thousands of new tasks. We adopt a variant of the \texttt{Craft} domain in Andreas \etal~\cite{andreas2017modular}. The \texttt{Craft} domain was originally developed for multi-task RL.

Safety-critical systems benefit from following constraints expressed as LTL
formulas~\cite{Alshiekh2018safe}. Shielded reinforcement learning prevents
agents from entering states which violate constraints that can be expressed in
LTL. This is critical for many real-world applications such as autonomous cars
that can adapt to new environments. The approach presented here is
complementary; it could provide an effective compositional shield that encodes
new constraints.
In the future, one might even be able to learn what constraints
are required to shield a system.
Several publications investigate combining Markov decision processes and
Q-learning with constraints specified in
LTL~\cite{Fu2014probably,Wen2015correct,Sadigh2014learning}.

Prior work has attempted to augment learning to satisfy constraints with
policies derived from those
constraints~\cite{Li2017reinforcement,Li2018policy}. This is a finer-grained
analysis of the formulas than we perform here. While this speeds up learning,
each formula requires significant training data.

\section{Model}\label{sec:model}

The model we introduce is presented in \cref{fig:model}.
It is inherently compositional, \ie the structure of the model reflects the
parsing structure of the input LTL formula.
Each LTL formula is parsed into a tree where the nodes are the operators or
predicates.
Operator and predicates are replaced by recurrent networks and are connected
with one another according to the parse tree of the LTL formula.
Every LTL formula generates a unique network that encodes the meaning of that
specific formula.

We employ an Advantage Actor-Critic,
A2C~\citep{sutton2000policy,mnih2016asynchronous}, agent.
Actor-critic models learn a value function, the critic that determines the score
of a state, and a policy, the actor which determines what to do at a given
state.
These could in principle be two separate networks.
The critic network evaluates state, $s$ at time $t$, $V_v(s_t)$ with learned
parameters $v$.
The actor network evaluates the effect of action, $a$, at time $t$, given a
state $s_t$, $A_\theta(s_t,a_t)$, with learned parameters $\theta$.
Advantage Actor-Critic models use the fact that the actor is computing an
advantage, a change in value between two states, to estimate the actor using the
value function of the critic.
Practically, this means that a single network is required from which both the
actor and the critic can be computed.
The compositional model shown in \cref{fig:model} is this shared network between
the actor and the critic.
The parameter updates of this network follow the standard methodology for
training recurrent networks with A2C presented in Minh \etal (2016)
\citep{mnih2016asynchronous}.

In prior work, when learning two different policies, the networks for those two
policies would be unrelated to one another.
The model presented here can be seen as a principled way to share weights
between these networks informed by the structure of LTL formulas.
All operators and symbols share weights both within a formula and between
updates, \ie there is only a single model for $\ltlnext$ or $gem$.

At training time, we supervise the agent with random LTL formulas.
Each formula is converted to a Buchi automaton using Spot~\cite{spot2}.
Note that we only consider and evaluate LTL with finite traces~\cite{de2013linear} here, meaning
that these automata are interpreted as having the semantics of finite automata.
Each training episode proceeds with the agent taking a sequence of actions.
Each action is evaluated against the automaton.
The predicates in the possible transitions of the automaton are evaluated.
If the predicates hold, the agent is given a small, 0.1, reward.
Staying in the same non-accepting state lowers the reward at a reward decay rate 0.8 for both \texttt{Symbol} and \texttt{Craft} domains.
If the predicates do not hold, the agent has violated the semantics of the LTL
formula and it receives a large negative, -1, reward.
If the predicates hold and the agent is in an accepting state, it receives a
large positive, 1, reward.
This reward structure encodes the notion that agents should follow an automaton
which encodes a particular LTL formula, although agents do not have access to
the automaton directly.
We employ curriculum learning to sort generated formulas and provide shorter
formulas first.
Short formulas are more likely to accept more strings and their shallower
corresponding models make error assignment more reliable.

The model is structured in such a way that knowledge flows between operators and
predicates; see \cref{fig:model}.
The current state of all operators is fed to children after being decoded by a
linear layer.
This allows models to communicate information about the current sequence of
steps to their children.
The next state of each child is passed to its parents after being decoded by a
linear layer.
This is the only form of communication between models.
Parents let their children know about the current state sequence and children
let the parent update their representations based on observations.
The next state of the root of the tree is decoded by the actor and the critic
and used to predict the next action.

We use the intuition developed in Kuo \etal (2020) \citep{kuo2020language} that
models which are composed of sub-networks can disentangle the meaning of words in a sentence without direct supervision.
In other words, the agent is never told what $\ltlnext$ is supposed to mean as
opposed to $gem$.
Yet this can be done automatically, since from the definition of LTL
formulas we understand that the computation required for every single operator
or predicate should be the same.
A learning method that attempts to find the most parsimonious explanation of the
meaning of these shared sub-networks should then hone in on their meaning as
components of LTL formulas by virtue of being forced to share computation when
the definition of LTL formulas demands it.
This intuition makes clear why the model presented here can recompose the sub-networks into new
formulas: it has the capacity to execute recurrent computations, it shares
computations in a way that the definition of LTL requires, and it is rewarded
when it replicates the computations that are required to satisfy LTL formulas.


\section{Experiments}

We demonstrate the model in two domains.\footnote{Source code is available at \href{https://github.com/ylkuo/ltl-zero-shot}{https://github.com/ylkuo/ltl-zero-shot}}
The first, \texttt{Symbol}, is designed to stress the symbolic reasoning abilities of RL agents.
An LTL formula is provided to an agent which must immediately produce a
satisfying assignment to that formula as a series of symbols.
The second, \texttt{Craft}, is designed to stress the multi-task execution capabilities
of RL agents in a robotic environment.
An LTL formula and a map containing a robot are provided to an agent which must
immediately find a sequence of moves that result in behavior of the robot that
satisfies the LTL command.

Four datasets are generated for each domain and each is tested against the
model and three ablations of the model.
A training dataset is first generated.
Note that even in the case of the training set, our method has a significant
advantage over prior work: our training set contains thousands to tens of
thousands of formulas which are all learned, as opposed to learning one or a
small handful of formulas.
Then three test sets are generated of increasing difficulty.
A test set which has roughly the same statistics as the training set in terms of
formula length, \ie 1 to 10 predicates with an average of 8, one which has 10-15
predicates, with an average of 13, and one which has 15-20 predicates with an
average of 18.
These test sets stress the generalization capabilities of the model and
demonstrate that even when faced with formulas that are well beyond any that
have been seen before in terms of complexity, the compositional nature of the
model often leads to correct executions.

\subsection{\texttt{Symbol} domain}
\label{sec:symbol}

The \texttt{Symbol} domain is introduced here as a new challenge for multi-task
LTL-capable agents.
It removes the map and focuses on generating accepting strings for an LTL
formula.
The map can be a crutch for agents, \eg crowded maps can have few paths making
even random exploration efficient.
The absence of particular resources on the map can also
significantly simplify the problem, \eg if there is no gem on the map, the agent
can't mistakenly pick up a gem.
In the \texttt{Symbol} domain, given a fixed inventory of symbols, the agent
predicts the assignment of symbols at each time step, a sequence of fixed length that will be accepted by
the LTL formula.

In the experiments reported here, we use an environment that has 5 symbols and
requires satisfying sequences of length 15.
Changing the number of symbols does make the problem more difficult but not
substantially so.
This is because the approach presented here separately learns networks for each
symbol making it robust to increasing the number of symbols; adding a symbol
corresponds to adding a single sub-network.

Data generation is complicated by the fact that generating random LTL formulas
produces uninteresting and easily satisfied instances. In a related domain, this
is a well-known property of random instances of boolean satisfiability, SAT,
problems~\citep{selman1996generating}. We adapt solutions from this community to
generating interesting collections of LTL formulas that are diverse and
difficult to satisfy.

The generation process begins by sampling a formula from a uniform distribution
over trees with one parameter: the prior distribution over the number of
elements (predicates and symbols). This formula is converted to a Buchi automata
using Spot~\cite{spot2}. For a fixed $n$, the number of strings up to length $n$
that the formula accepts is computed using a dynamic programming algorithm. This
recursively computes the number of accepted strings for each time step, $N$. All
formulas which have an acceptance ratio of more than 0.0001\% are rejected as
being too easy. Most randomly generated formulas tend to accept almost all
strings. This provides hardness but does not guarantee variety.

In a second step, a second formula is generated as a function of the first
formula. This uses a random process that picks a random subtree of the formula
and replaces it with a new, also random, subtree. If the second formula is
rejected, the process restarted based on the acceptance criteria
above. If accepted, random accepted strings are sampled from the first formula
and verified against the second. If more than 10\% of strings that satisfy the
first formula also satisfy the second, it is rejected as not diverse enough.

This may appear to be a rather laborious process, but without ensuring both
hardness and diversity, we found that the generated formulas are uninteresting
and fairly homogeneous in what strings or robot moves they accept. This is
evidenced by the fact that without this process, there often existed a single
string or sequence of robot moves which was accepted by the majority of
generated formulas. After this more complex generation process, the resulting
dataset is far more challenging, as is reflected in the experiments section by
the low performance of baseline methods. The statistics of the generated
datasets are shown in \cref{fig:formulas}. The generation process is heavily
biased against short formulas as they tend to be overly permissive and overlap
with one another in meaning.



\subsection{\texttt{Craft} domain}
\label{sec:minecraft}

\begin{figure}
    \centering
    \includegraphics[width=0.4\textwidth]{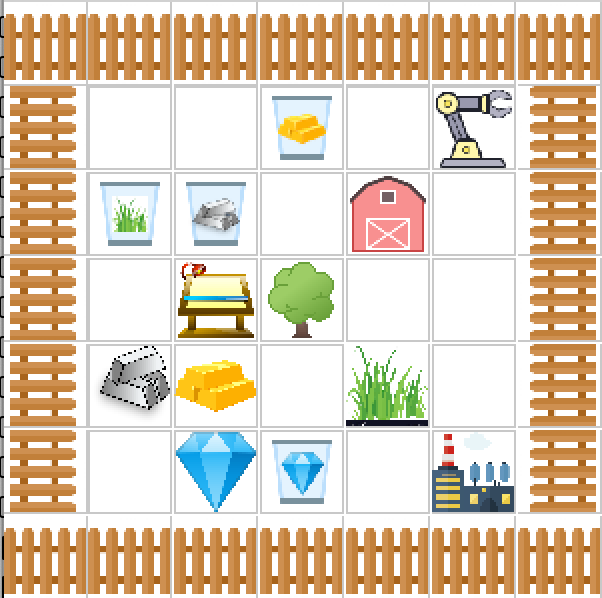}
    \caption{A $7 \times 7$ \texttt{Craft} domain example showing all the various elements
      available in the domain. The robot is in the top right corner and is
      controlled by the RL agent presented here. Four trash cans contain
      different resources: gold, grass, silver, and gems. A work bench, a
      tree, a tool shed, and a factory provide additional destinations. The
      general intent is to model a Minecraft-like environment in which resources
      are processed through several stages.}
    \label{fig:craft_world}
\end{figure}

We use the Minecraft-like world already used by several prior multi-task
learning reinforcement learning publications
\cite{andreas2017modular,Toro2018teaching}.
See~\ref{fig:craft_world} for an example map that contains all of the elements
of the \texttt{Craft} domain.
It mixes together unmovable objects (trees, tool sheds, workbenches, and
factories) with resources that can be manipulated (silver, grass, gold, and gems).
We generate random \texttt{Craft} maps with random initial positions for the robot arm.
The environment is 4-connected with an additional ``use'' action for the arm
which picks up or drops the resource.
The overall intent is that resources can be manipulated through multiple
processing stages resulting in a range of interesting tasks.

We extend this environment with a recycling bin so that agents can discard
items.
This allows the agents to satisfy formulas that require owning an item for only
part of the action.

Unlike in the \texttt{Symbol} domain, where most random formulas were trivially
satisfied without our procedure to find hard and diverse LTL instances, in the
\texttt{Craft} domain, most formulas are trivially unsatisfiable.
Constraints on where the robot is at any one time and the fact that it must
traverse the map step by step, make even basic formulas such as
$\ltlglobally\textsc{gem}$, the gem must always be held, unsatisfiable in all
but the most fortunate situations.
If the robot happens to be next to the gem, this formula can be satisfied by
picking the gem up.
If the robot is not next to the gem, no one action the robot can take will
satisfy this formula.

We could employ rejection sampling and resample formulas until they can be
satisfied.
This would be wasteful and produce the same results as the following
transformation.
Since formulas are unsatisfiable in \texttt{Craft} largely because the robot
does not have time to satisfy them, we introduce a transformation that gives the
robot time.
Whenever the robot needs to satisfy a predicate, we transform that predicate into
``\texttt{closer}(\emph{predicate}) $\bigcup$ \emph{predicate}'', which states
that the robot must get closer to its goal until it is reached.
To make following this command feasible, the robot is given an additional feature
which is the Manhattan distance to the predicate.
This does not otherwise change the semantics of any of the formulas, nor does it
make zero-shot generalization easier as the robot must still understand the
command as a whole.
The transformation also does not count toward the formula lengths we produce; it
merely gives the robot some time.
Indeed, this transformation actually makes learning harder because the agent
does not receive any meaningful feedback if it runs out of time.
It has no way of knowing if an episode failed because it could not reach a goal
in time or because some constraint was violated.
In the future, we intend to investigate how to make the causes of failure more
transparent to the agents, thereby hopefully speeding up learning.






\subsection{Generated formulas}
\label{sec:formula-generation}

\begin{figure*}[th!]
  \centering
  \begin{tabular}{lcccc}
    Formula set & symbols & tree nodes & tree depth & automata states \\[0.5ex]
    \texttt{Symbol} (train) & $3.28 \pm 0.49$ & $\phantom{0}9.14 \pm 1.16$ & $4.49 \pm 0.96$ & $4.21 \pm 0.52$ \\
    \texttt{Symbol} (test, 1-10) & $3.36 \pm 0.50$ & $\phantom{0}9.18 \pm 1.13$ & $4.70 \pm 0.94$ & $4.20 \pm 0.45$ \\
    \texttt{Symbol} (test, 10-15) & $4.64 \pm 0.77$ & $13.05 \pm 1.49$ & $6.04 \pm 1.14$ & $5.32 \pm 1.75$ \\
    \texttt{Symbol} (test, 15-20) & $6.11\pm 0.73$ & $17.99 \pm 1.42$ & $8.05 \pm 1.56$ & $6.69 \pm 3.44$ \\[0.5ex]
    \texttt{Craft} (train) & $2.76 \pm 0.63$ & $\phantom{0}7.94 \pm 1.49$ & $4.35 \pm 1.04$ & $3.40 \pm 0.77$ \\
    \texttt{Craft} (test, 1-10) & $2.78 \pm 0.64$ & $\phantom{0}7.83 \pm 1.46$ & $4.22 \pm 0.98$ & $3.37 \pm 0.74$ \\
    \texttt{Craft} (test, 10-15) & $4.33 \pm 0.82$ & $12.94 \pm 1.54$ & $6.62 \pm 1.05$ & $3.73 \pm 0.96$ \\
    \texttt{Craft} (test, 15-20) & $6.06 \pm 0.75$ & $18.14 \pm 1.31$ & $7.94 \pm 1.37$ & $4.44 \pm 1.96$ \\
  \end{tabular}
  \caption{Statistics of the generated formulas used in each of the two kinds of
    experiments, on the \texttt{Symbol} domain and the \texttt{Craft} domain. The training set is
    generated with the same mechanism as (test, 1-10), \ie, containing formulas
    that have been 1 and 10 elements. Training and test sets are randomly
    generated but verified to be disjoint. Two additional more difficult test
    sets are created, 10-15 and 15-20 which have longer formulas with, 10 to 15
    elements and 15 to 20 elements, respectively. Each test set is summarized
    using number and standard deviation of the number of symbols in the formula,
    total tree nodes in the parse of the formula, depth of the parse of the
    formula, and the states in the Buchi automata for those formulas. Note that
    while formulas are generated randomly, very short formulas are unlikely to
    appear; see~\protect\cref{sec:symbol} and~\protect\cref{sec:minecraft} for a
    discussion on how to generate difficult random collections of LTL formulas
    in each of the two domains.}
  \label{fig:formulas}
\end{figure*}

Four sets of formulas are generated for the experiments in the \texttt{Symbol}
and the \texttt{Craft} domains; see \cref{fig:formulas}. The training set for
each has between 1 and 10 elements, \ie operators and symbols. An out of domain
test set has the same number of elements but does not overlap; this tests
zero-shot execution. To further demonstrate that this approach generalizes, we
produce two additional test sets of even longer formulas than those seen in the
training set. Note that these latter test sets are both longer and more
difficult than the training sets.

%
%

\subsection{Hyper-parameters}

We train the Advantage Actor Critic, A2C, model with
RMSprop~\citep{tieleman2012rmsprop} optimizer for both domains.
Agents must explore the space of symbols and moves thoroughly since our data
generation process ensures that the LTL formulas are strict and few operations
resulting in accepting states.
To encourage this, we set a higher entropy weight of 0.1 and set the reward
decay, $\gamma$, to 0.9 during training.
While training, the next move is an action from a distribution over possible
actions.
While testing, a deterministic policy chooses the optimal move.
We observed a significant performance drop if the agents chose deterministically
at training time or stochastically at test time.

Each sub-network, which represents an operator or predicate, is represented
using a gated recurrent unit, GRU \citep{cho2014properties}.
The hidden sizes are 64 for both domains.
The \texttt{Symbol} domain has only one hidden layer and the 
\texttt{Craft} domain has two hidden layers since it is significantly more complex.
A single linear layer per sub-network is always used as the decoder that
transmits information from children to parents and a single linear layer
transmits information from parents to children in the model structure.

We consider only LTL formulas with finite traces~\cite{de2013linear} here, although nothing in our method
prevents executing non-finite LTL.
The interpretation of the results for LTL without a finite horizon is
considerably more complex and requires new metrics because it must consider when
a model fails, not just if it fails.

We perform 15 rollouts at every iteration.
If at the end of 15 steps the model is not in an accepting state, it is
considered to have failed and given zero reward as if it had taken
an action that violated the semantics of the LTL formula.

\subsection{Baselines}

In addition to our model, we test three baselines.
The ``no structure, no language'' baseline takes the current state of the world
and attempts to predict what action the agent should take next.
It is a standard recurrent A2C agent that does not observe the LTL formula at
all.
This is the performance an off-the-shelf agent would have knowing only the
environment.
The ``no structure'' baseline takes as input the LTL formula, learns an
embedding of that formula into a single 32-dimensional vector, and then
attempts to follow it.
To create the embeddings, each operator and predicate in the formula is encoded
as a one-hot vector.
Sequences of these one-hot vectors are passed to an RNN which is co-trained to
produce the embedding of the formula, much like the CNN produces an embedding of
the environment.
The formula is provided to every sub-network just like the observations of the
environment.
This is the performance an off-the-shelf agent would have that does not
understand compositionality.
The ``no time'' baseline is an ablation of our model; it is structured but is
missing any recurrent connections, \ie all symbols are feed-forward networks
instead of GRUs.
This is the performance an off-the-shelf non-recurrent A2C agent would have
because it cannot keep track of its progress through the formula.

\subsection{Results}

\begin{figure*}[th!]
  \vspace{0.7ex}
  \centering
  \begin{tabular}{lcccc}
    & In domain & Out of domain (1-10) & Out of domain (10-15) & Out of domain (15-20) \\
    \hline
    \texttt{Symbol} domain \\
    no structure, no language & 0.12 & 0.15 & 0.10 & 0.12 \\
    no structure & 0.65 & 0.74 & 0.39 & 0.23 \\
    no time & 0.01 & 0.01 & 0.02 & 0.04 \\
    ours & 0.97 & 0.90 & 0.60 & 0.39 \\
    \hline
    \texttt{Craft} domain\\
    no structure, no language & 0.43 & 0.48 & 0.38 & 0.31 \\
    no structure & 0.54 & 0.59 & 0.36 & 0.31 \\
    no time & 0.43 & 0.31 & 0.19 & 0.19 \\
    ours & 0.73 & 0.67 & 0.52 & 0.57 \\
  \end{tabular}
  \caption{The performance of our model finding satisfying actions for LTL
    formulas. Formulas which are perfectly executed are reported as successes,
    any errors are considered a failure. Each formula set contains 10,000
    formulas for the \texttt{Symbol} domain, and 4,000 formulas for the
    \texttt{Craft} domain. ``In domain'' refers to training and testing on the
    same formulas. Note that this task is already far more complex than what
    existing methods can do as the model must learn to execute thousands of LTL
    formulas given the formula as input. We then test out of domain, to
    demonstrate that our model can zero-shot execute new formulas. Finally, two
    increasingly complex out of domain scenarios are tested. One in which has
    formulas that are 1.5 times as long, and one with formulas that are twice as
    long. Note the poor performance of the baseline methods and ablations; our
    method performs far better.}
  \label{fig:results-table}
\end{figure*}

\begin{figure*}[th!]
  \vspace{0.7ex}
  \centering
  \begin{tabular}{cccc}
    \includegraphics[width=0.22\textwidth]{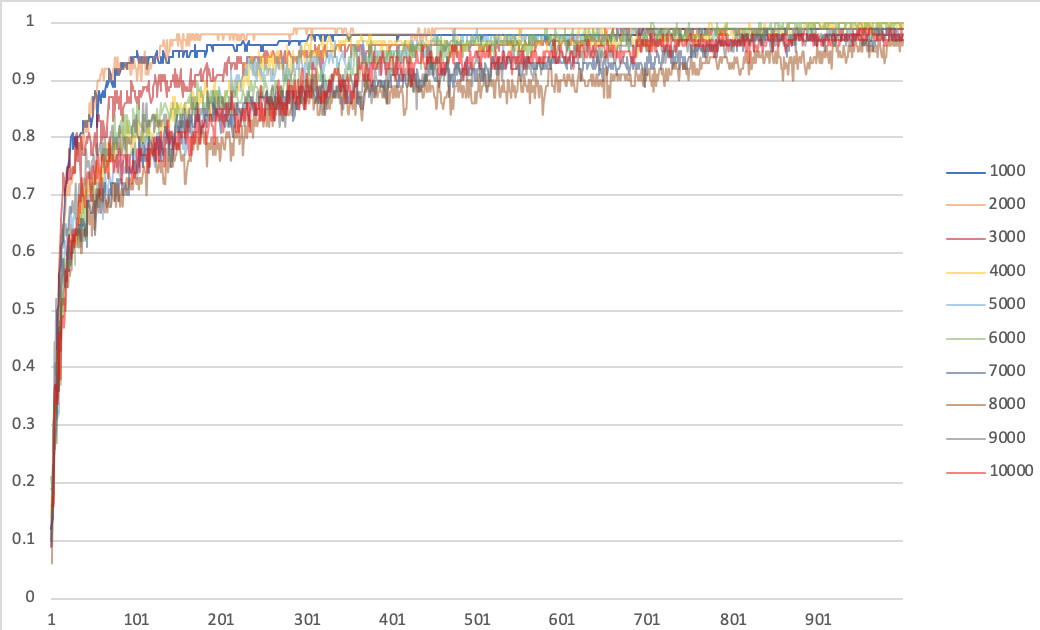}
    &\includegraphics[width=0.22\textwidth]{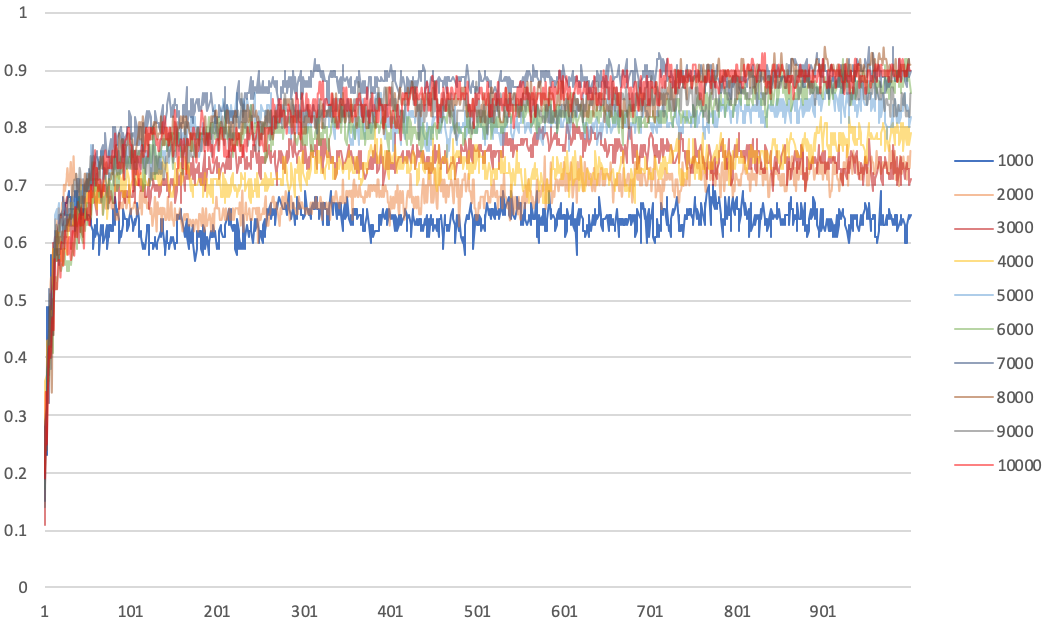}
    &\includegraphics[width=0.22\textwidth]{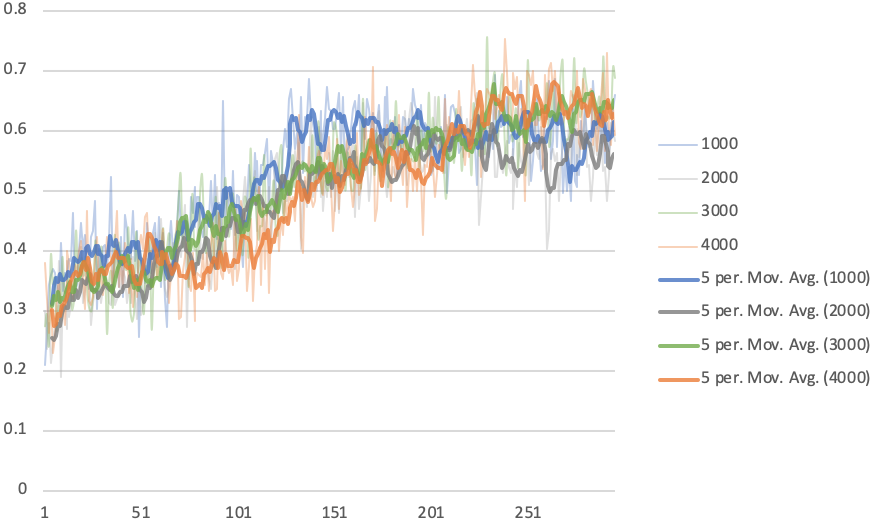}
    &\includegraphics[width=0.22\textwidth]{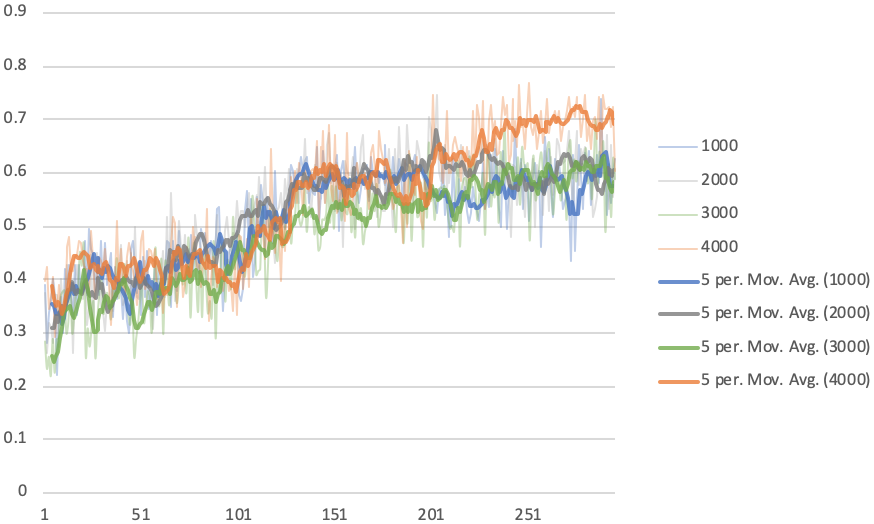}\\
    \texttt{Symbol} in domain &
    \texttt{Symbol} out of domain (1-10) &
    \texttt{Craft} in domain &
    \texttt{Craft} out of domain (1-10)
  \end{tabular}
  \caption{Performance of the model as the number of training and test formulas
    increases. The x-axis is the number of model updates performed, each unit is
    200 updates for \texttt{Craft} and 500 updates for \texttt{Symbol}. The \texttt{Symbol} domain on {\blue 1,000} and
    {\red 10,000} formulas can reach similar performance but a 25\% performance difference for out of domain (65\% \vs 90\% accuracy). The \texttt{Craft} domain on {\blue 1,000} to {\orange
      4,000} formulas shows a similar performance improvement. Seeing on the
    order of 10,000 formulas allows good generalization to new formulas. Note
    that any minor error made while executing the formulas was considered a
    failure.}
  \label{fig:results-graphs}
\end{figure*}

The performance of the \texttt{Symbol} domain is reported on formula sets of
10,000 formulas each, while the \texttt{Craft} domain, which is considerably
slower, is reported on 4,000 formulas each; see \cref{fig:results-table}.
In domain, a difficult task where one of 10,000 LTL formulas that have been seen
before is provided and must be executed correctly, has 97\% accuracy for
\texttt{Symbol} and 73\% accuracy for \texttt{Craft}.
Ablations of our method show that without the particular compositional structure
imposed the performance is two-third for ``no structure'' and even lower
for the other ablations.
Every part of our model is critical to performance as ablating any part away
hurts performance tremendously.
No previous method can learn to execute such formulas.

The performance on formulas that have similar statistics to the ones in the
training degrades only slightly when testing on new formulas, and is shown in the
second column of \cref{fig:results-table}, ``Out of domain (1-10)''.
This shows zero-shot generalization and that in all cases our method generalizes
well.

Absolute performance is a function of how many formulas our model is trained
with.
Several thousand formulas must be seen before generalization to new formulas and
longer formulas is reliable, see \cref{fig:results-graphs}.
In the \texttt{Symbol} domain, 1,000 training formulas result in 65\% accuracy
out of domain while 10,000 training formulas result in 90\% out of domain
accuracy.
As the number of training formulas increases, generalization improves.
%
%
%
To extrapolate how many formulas would be needed for a given level of
performance, we used a least squares fit to a logarithmic function, which had
$R^2$ of $0.99$.
This predicts that performance would approach 100\% at around 24,000 formulas.
In other words, seeing 24,000 formulas that contain up to 10 predicates and
operators is likely to be enough to perfectly generalize to all formulas of that
length.
To put this into context, with 5 symbols in the \texttt{Symbol} domain presented
here, there are approximately $10^{10}$ formulas of length 10.
This means that our
network sees only one-millionth of all possible formulas before generalizing,
not including its ability to generalize to longer formulas.
Similarly in the \texttt{Craft} domain, 1,000 training formulas result in 59\%
accuracy out of domain, while 4,000 training formulas result in 70\% accuracy
out of domain.
Together, these results show that our method learns to generalize formulas and to
execute them zero-shot.




\section{Conclusion} 

We created a principled network that is able to encode the dynamics required to solve
LTL formulas in a compositional manner.
This represents a new and powerful type of multi-task learning, where learning
occurs on one set of tasks and generalizes to all others.
Coupled with a semantic parser, this model could execute linguistic commands that
refer to temporal relations; we intend to pursue this in the future.
While we believe that such network architectures are useful for other logics,
such as first-order logic, how precisely they should be extended to even more
exotic modal logics or second-order logics is unclear.
We would additionally like to encompass more powerful logics like CTL$^{*}$.
Theoretically characterizing the structures required in neural networks to
generalize out of domain to new problems generated from particular logics is a
new and interesting problem that may impact our understanding of other types of
generalization.

Perfect, 100\% generalization, performance on both \texttt{Symbol} and
\texttt{Craft} appears to be within reach, although computationally this remains
rather expensive.
Around 10 to 100 times more computationally intensive than the results reported
here.
We intend to investigate what additional priors, curricula, or training
algorithms can speed up learning to saturate performance in a more effective
time frame.
Overall, depending on the experiment and number of formulas, the results here took between 13 hours to 3 days to generate with each run executing on
an Nvidia Titan X.
Note that while relatively computationally intensive, such networks need only be
trained once for any domain; due to their ability to zero-shot generalize to new
formulas.

As it stands, the reward function designed here to supervise the RL agent is
powerful but lacks some critical feedback.
For example, the agents do not know what went wrong.
Some feedback about the constraint that was violated could localize errors
within particular sub-networks.
This would be akin to telling someone that they had picked up gold instead of
silver; clearly very useful information.
Providing such targeted feedback to a compositional network seems possible in
theory, one would need to more carefully determine which sub-networks could be
at fault and emphasize updates to those sub-networks for a failed trial.
Turning this intuition into a practical learning mechanism remains an open
problem.

In the future, we intend to combine this work with that presented by Kuo \etal
2020 \citep{kuo2020language} and create a single agent capable of following
complex linguistic commands in continuous environments.

\bibliographystyle{IEEEtran}
\renewcommand*{\bibfont}{\footnotesize}
\bibliography{references}

\end{document}